\documentclass[10pt]{article}

\usepackage{tikz}
\usetikzlibrary{calc,intersections,through,backgrounds}

\usepackage[noadjust]{cite}

\usepackage{float}

\usepackage{amsmath}
\usepackage{amsfonts}

\usepackage{url}
\usepackage{hyperref}
	\hypersetup{
    colorlinks=true,
    linkcolor=blue,
    citecolor=blue,
    filecolor=magenta,      
    urlcolor=cyan,
	}

\usepackage{makecell}
\usepackage[font={small,it}]{caption}

\begin{document}

\title{Reservoir Computing Hardware with\\ Cellular Automata}

\author{Alejandro Mor\'an,  Christiam F. Frasser  and  Josep L. Rossell\'o}
\date{%
\small Electronic Engineering Group, Physics Department, Universitat de les Illes Balears, Spain.\\
E-mail: j.rossello@uib.es\\
    \today
}






\maketitle
\begin{abstract}
\noindent Elementary cellular automata (ECA) is a widely studied one-dimensional processing methodology where the successive iteration of the automaton may lead to the recreation of a rich pattern dynamic. Recently, cellular automata have been proposed as a feasible way to implement Reservoir Computing (RC) systems in which the automata rule is fixed and the training is performed using a linear regression. In this work we perform an exhaustive study of the performance of the different ECA rules when applied to pattern recognition of time-independent input signals using a RC scheme. Once the different ECA rules have been tested, the most accurate one (rule 90) is selected to implement a digital circuit. Rule 90 is easily reproduced using a reduced set of XOR gates and shift-registers, thus representing a high-performance alternative for RC hardware implementation in terms of processing time, circuit area, power dissipation and system accuracy. The model (both in software and its hardware implementation) has been tested using a pattern recognition task of handwritten numbers (the MNIST database) for which we obtained competitive results in terms of accuracy, speed and power dissipation. The proposed model can be considered to be a low-cost method to implement fast pattern recognition digital circuits.
\end{abstract}


%

\section{Introduction}\label{sec:introduction}

Reservoir Computing (RC) \cite{maass2002, jaeger2009} is an attractive machine learning alternative due to its simplicity and computationally inexpensive learning process. In the standard RC approach the input is connected to a randomly initialized Recurrent Neural Network (RNN) and the training process is only applied to the output layer weights using linear or logistic regression. The RC learning methodology has been considered to be a practical alternative to the backpropagation through time (BPTT) algorithm. In the special case of processing time-independent signals, the use of RNN is normally providing results with lower precision due to their long short-term memory (LSTM) properties \cite{Hochreiter1997}. For these cases the feed forward neural network architecture (FFNN) is normally used but with a cost of a more complex training methodology if compared with the linear regression that is normally applied to RC.  RC has been successfully applied in numerous domains, such as robot control \cite{Antonelo2015}, image/video processing \cite{Jalalvand2015}, wireless sensor networks \cite{Bacciu2014}, financial forecasting \cite{Lin2009} or to the monitoring of physiological signals \cite{Buteneers2013}. In this context, fast and efficient hardware designs implementing RC systems can be an interesting alternative for many of these applications which require a real-time and intensive data processing and/or the use of low-power devices to ensure a long battery lifetime.

\begin{figure}
 	\centering
 	\includegraphics[width=\columnwidth]{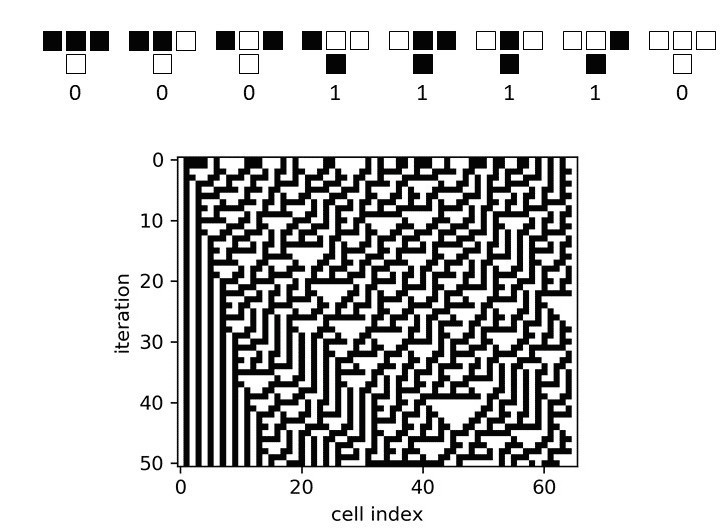}
 	\caption{Basic mechanism of elementary cellular automata (rule 30) showing its temporal evolution from random initial conditions and fixed boundary conditions.}
 	\label{fig:ecaRule30}
\end{figure}

Elementary cellular automata (ECA) is the simplest class of 1-dimensional cellullar automata \cite{wolfram2002new} that is found to provide a rich and complex dynamic behavior that is also reproduced by more complex CA schemes. ECA systems consists of a 1D string of cells that can be settled in two possible states (high or low) and evolve in discrete time steps guided by the interaction between nearest neighbors. Since there are a total of $2^3=8$ binary states for three cells (i.e. a given cell and its nearest neighbours), there are a total of $2^8=256$ possible rules of interaction, which are labeled from 0 to 255 following a standard convention. This convention is illustrated in Fig.\ref{fig:ecaRule30}, in which the interaction rule 30 is shown as an example of application.  

Different image analysis and  processing methods have been developed based on CA, e.g. \cite{haralick1987image, ge2012vision, dyer1981parallel,wang2017stagewise}. This is the case of the multiple attractor cellular automata (MACA) that has been used to classify 1-dimensional arrays of binary data \cite{ganguly2002evolving}, working as an associative memory and where the processing is restricted to limit cycle dynamics. Other application example in this line is two-dimensional clustering  \cite{tzionas1994new,chady1997evolution,maji2003theory}, which also relies on the associative memory approach using MACA.

In this work an efficient hardware implementation of a reservoir computing system based on cellular automata (ReCA) is proposed. ReCA systems differ from more traditional associative memory approaches in the inclusion of the transient and chaotic temporal evolution of the automata. Therefore, the dynamic used is not limited to fixed points and limit cycles but extended to transient and chaotic behaviors. Following the developments performed in \cite{yilmaz2015connectionist, Nichele2017deep, Nichele2017nonuniform, McDonald2017}, the temporal evolution of a cellular automata (CA) (both in the transient and stationary states) is used as a reservoir.  In the proposed ReCA system the CA states are readout by an output layer and the cellular automata's initial state is equal to the input signal to be processed. Firstly, we develop an exhaustive study of the accuracy of different ECA rules when applied to pattern recognition tasks. The most accurate rule that is finally selected (rule 90) can be implemented using a simple digital circuitry.  Although this is not the first time in which a cellular automata is implemented in digital circuitry for research purposes (see Ref. \cite{Halbach2004}), to the best of our knowledge this is the first time in which a full ReCA circuitry oriented for pattern recognition is implemented and tested in hardware. The proposed circuitry is synthesized in a medium-sized Field Programmable Gate Array (FPGA) and tested to processing the MNIST handwritten numbers dataset \cite{lecun1998gradient}. The proposed model is compared with previously-published hardware implementations oriented to pattern recognition showing highly competitive results.

 

\section{Methods}\label{sec:methods}
\subsection{Reservoir Computing principles}
\begin{figure}
\centering
\begin{tikzpicture}
[inner sep=1mm,place/.style={circle,draw=blue!50,fill=blue!20,thick},
sortida/.style={circle,draw=black!50,fill=black!20,thick},entrada/.style={circle,draw=red!50,fill=red!20,thick}]
\node (il)  at (-1.6,0) {Input layer};
\node (ol)  at (4.7,0) {Output layer};
\node (v2)  at (1.6,-0.5) {Reservoir};
\node[place] (w11)  at (0,0) {};
\node[place] (w12) at (0,1) {};
\node[place] (w13) at (0,2) {};
\node[place] (w14) at (0,3) {};
\node[place] (w21)  at (1,0) {};
\node[place] (w22) at (1,1) {};
\node[place] (w23) at (1,2) {};
\node[place] (w24) at (1,3) {};
\node[place] (w31)  at (2,0) {};
\node[place] (w32) at (2,1) {};
\node[place] (w33) at (2,2) {};
\node[place] (w34) at (2,3) {};
\node[place] (w41)  at (3,0) {};
\node[place] (w42) at (3,1) {};
\node[place] (w43) at (3,2) {};
\node[place] (w44) at (3,3) {};
\node[sortida] (s1) at (4.5,0.5) {};
\node[sortida] (s2) at (4.5,1.5) {};
\node[sortida] (s3) at (4.5,2.5) {};
\node[entrada] (e1) at (-1.5,0.5) {};
\node[entrada] (e2) at (-1.5,1.5) {};
\node[entrada] (e3) at (-1.5,2.5) {};
\draw [->,dashed] (w44) -- (s3);
\draw [->,dashed] (w14) -- (s3);
\draw [->,dashed] (w43) -- (s3);
\draw [->,dashed] (w13) -- (s3);
\draw [->,dashed] (w42) -- (s3);
\draw [->,dashed] (w12) -- (s3);
\draw [->,dashed] (w41) -- (s3);
\draw [->,dashed] (w11) -- (s3);
\draw [->,dashed] (w44) -- (s2);
\draw [->,dashed] (w14) -- (s2);
\draw [->,dashed] (w43) -- (s2);
\draw [->,dashed] (w13) -- (s2);
\draw [->,dashed] (w42) -- (s2);
\draw [->,dashed] (w12) -- (s2);
\draw [->,dashed] (w41) -- (s2);
\draw [->,dashed] (w11) -- (s2);
\draw [->,dashed] (w44) -- (s1);
\draw [->,dashed] (w14) -- (s1);
\draw [->,dashed] (w43) -- (s1);
\draw [->,dashed] (w13) -- (s1);
\draw [->,dashed] (w42) -- (s1);
\draw [->,dashed] (w12) -- (s1);
\draw [->,dashed] (w41) -- (s1);
\draw [->,dashed] (w11) -- (s1);

\draw [<-,red] (w44) -- (e3);
\draw [<-,red] (w33) -- (e3);
\draw [<-,red] (w41) -- (e3);
\draw [<-,red] (w44) -- (e2);
\draw [<-,red] (w22) -- (e2);
\draw [<-,red] (w21) -- (e2);
\draw [<-,red] (w23) -- (e1);
\draw [<-,red] (w11) -- (e1);

\draw [->] (w11) -- (w23);
\draw [->] (w21) -- (w12);
\draw [->] (w21) -- (w32);
\draw [->] (w31) -- (w23);
\draw [->] (w41) --(w31);
\draw [->] (w12) -- (w13);
\draw [->] (w22) -- (w34);
\draw [->] (w32) -- (w24);
\draw [->] (w42) -- (w21);
\draw [->] (w13) -- (w34);
\draw [->] (w23) -- (w14);
\draw [->] (w33) -- (w21);
\draw [->] (w43) -- (w14);
\draw [->] (w44) -- (w33);
\draw [->] (w44) -- (w31);
\draw [->] (w34) -- (w42);
\end{tikzpicture}
\caption{Reservoir computing scheme. Three blocks: Input layer, reservoir and the output layer compose the circuitry. }
\label{reservoir}
\end{figure}
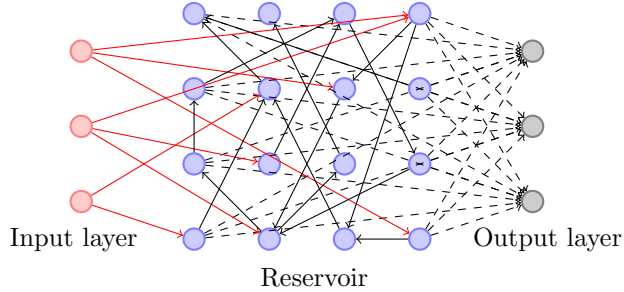

In the Reservoir Computing architecture, the internal weights between the computational nodes of the reservoir are kept fixed and only the connections to a non-recurrent output layer (the readout) are modified by learning (see Fig.\ref{reservoir}). This strategic design avoids the use of  complex back-propagation  methods, thus reducing the training to a classical linear regression problem. Depending on the typology of the computational nodes, the RC systems can be defined as Echo-State Networks, Liquid State Machines or ReCA (when using classical artificial neurons, spiking neurons or CA respectively).

The architecture of a reservoir computing system therefore consists of a total of $N$ internal processing nodes, each one providing a given value $x_i(k)$ where $i\in \lbrace 1, 2,..,N\rbrace$, $N$ is the number of computational elements in the reservoir, $k$ represents the evolution during time ($k\in \lbrace 1, 2, ...,M\rbrace$) and $M$ is the number of selected iterations.  The reservoir state [$\boldsymbol{x(k)}=\left( x_1, x_2, ... x_N\right)$] is therefore updated during $M$ time steps according to a nonlinear function:  

\begin{equation}
\boldsymbol{x(k)}=F\big[\boldsymbol{x(k-1), \boldsymbol{u(k-1)}}\big]
\label{dynamics}
\end{equation}

That is, the reservoir state depends on the input ($\boldsymbol{u(t)}$) and the previous state. The main objective of the reservoir is to correlate the actual input with $Q$ categories to be discovered. For this purpose, and taking a total of $N\cdot M$ descriptors ($N$ computational nodes evolved during $M$ time steps) the output layer of the reservoir will provide a total of $Q$ values [$\boldsymbol{\hat{y}}=\left( \hat{y}_1, \hat{y}_2, ..., \hat{y}_Q \right)$], each output representing a given category to be recognized in the case of pattern recognition or a variable to be predicted for regression. The output layer generate the estimation $\boldsymbol{\hat{y}}$ by applying a linear combination of the reservoir states during the $M$ time steps:

\begin{equation}
\boldsymbol{\hat{y}}= \boldsymbol{x}\boldsymbol{W}
\label{eq2}
\end{equation}

\noindent where $\boldsymbol{x}$ is a time-independent vector composed by $M\cdot N$ parameters (all the $x_i(k)$ valures considered during the $M$ time steps), and parameter $\boldsymbol{W}$ is a proper $M\cdot N \times Q$ weight matrix. To estimate $\boldsymbol{W}$, different ($L$) training samples must be considered to adjust these coefficients with respect to  $Q$ known outputs. Therefore, an $L\times Q$ matrix $\boldsymbol{Y}$ is needed as the ground truth. The $qth$ column ($\boldsymbol{Y^q}$) is composed by a total of $L$ samples taken as training set. For each sample, $\boldsymbol{\hat{y}}$ is the approximation provided by the reservoir activity so that the error with respect to the expected value $\boldsymbol{Y_i}$ ($ith$ row of matrix $\boldsymbol{Y}$) must be minimized. One of the most simple ways to derive $\boldsymbol{W}$ is through the Moore-Penrose pseudoinverse of $\boldsymbol{X}$ applied to $\boldsymbol{Y}$. Defining the design matrix $\boldsymbol{X}$ as the $M\cdot N$ descriptors coming from the reservoir for all the $L $ samples (an $L \times M\cdot N$ matrix), the weights $\boldsymbol{W}$ may be obtained as:

\begin{equation}
\boldsymbol{W}=\left(\boldsymbol{X^T}\boldsymbol{X} \right)^{-1}\boldsymbol{X^T}\boldsymbol{Y}
\label{pesos}
\end{equation}

Equation (\ref{pesos}) is a simple way to obtain the key coefficients $\boldsymbol{W}$. Nevertheless in this work we use a multinomial logistic regression or maximum entropy classifier that present better results than the Moore-Penrose method. The maximum entropy classifier is a linear model that generalizes the logistic regression method \cite{hosmer2013applied} to multiclass classification problems, and is explained in the next subsection.

\subsection{Softmax regression}  
\label{subsec:logistic}
	Softmax regression \cite{kwak2002multinomial}, also known as multinomial logistic regression or maximum entropy classifier, is a linear model that generalizes the logistic regression method \cite{hosmer2013applied} to classify multiple categories.
	
We define the Softmax function $S$ as:
	\begin{equation}\label{eq:softmax_function}
	S(\boldsymbol{\hat{y}}_q) = \frac{\exp{(\boldsymbol{\hat{y}}_q)}}{\sum_{q'} \exp{(\boldsymbol{\hat{y}}_{q'}})},
	\end{equation} 
which is computed for all samples and categories. From \eqref{eq:softmax_function} and $\boldsymbol{\hat{Y}}$, we construct the matrix $S(\boldsymbol{\hat{Y}})$ with information of each sample $i$ to belong to a certain category $q$.
	
	Given the design matrix $\boldsymbol{X}$, the $ith$ sample is denoted as $\boldsymbol{X}_i$, $i=1,\dots,L$ and for the corresponding one-hot enconded labels we use $\boldsymbol{L}_{(i)}$. The outputs finally selected are the most likely categories, $\textnormal{argmax}\left[ S(\boldsymbol{\hat{Y}})\right]$, or the one-hot encoded output $\boldsymbol{\hat{L}}=\textnormal{onehot}\Big\{ \textnormal{argmax}\left[ S(\boldsymbol{\hat{Y}})\right] \Big\}$.   
	
	To explain the softmax regression, lets first consider we have only one training example ($L=1$). The first step is to compute the logit vector $\boldsymbol{\hat{y}}$ using equation \eqref{eq2}. Next, the logit vector is converted to probabilities via a Softmax function \eqref{eq:softmax_function}. The model is trained by minimizing the cross-entropy between $S(\boldsymbol{\hat{y}})$ and the known one-hot encoded label $\boldsymbol{l}$:
	\begin{equation}\label{eq:crossent}
	\mathcal{D}\left( S(\boldsymbol{\hat{y}}),\boldsymbol{l}\right) = - \sum_{q}\boldsymbol{l}_q \log{S(\boldsymbol{\boldsymbol{\hat{y}}}_q)},
	\end{equation}
	
	where $\boldsymbol{l}_q$ is one or zero depending whether the index $q$ matches the corresponding category or not. Equation \eqref{eq:crossent} measures the Kullback-Leibler divergence between the probability vector $S(\boldsymbol{\hat{y}})$ and the actual one-hot encoded label (similar to a distance between probability distribution functions). For the general case in which a total of $L$ samples are considered, we minimize the average cross-entropy, which  is often referred to as the cross-entropy loss. Generalizing to $L>1$, a more general form of the loss function is
	\begin{equation}\label{eq:lossfunction}
	\mathcal{L}\left(S(\boldsymbol{\hat{Y}}),\boldsymbol{L}\right) = \frac{1}{L}\sum_{i=1}^{L}\mathcal{D}\left( S(\boldsymbol{\hat{Y}}_{i}),\boldsymbol{L}_{i}\right) + \frac{1}{C} \vert\vert \boldsymbol{W}\vert\vert_F,
	\end{equation}	
	where the first term is the average cross-entropy and the second is a regularization term where $\vert\vert \cdot \vert\vert_F$ denotes the Frobenius norm \cite{trefethen1997numerical}. Notice that a small value of the regularization parameter ($C$) penalizes large weights to avoid overfitting. This loss function can be minimized using gradient descent or more sophisticated optimization methods. In our case, we used the limited memory BFGS algorithm \cite{liu1989limited} provided by the Scikit-learn Python module \cite{pedregosa2011scikit} using double precision floating point numbers. However, for the hardware implementation we trained the model using the Adam stochastic optimization method \cite{kingma2014adam} and direct (fake) trained quantization in 8 bits to decrese the number of logic elements and computational power.
	
\begin{figure}[ht!]
\centering
\begin{tikzpicture}
[scale=0.7,reservoir/.style={rectangle,draw=blue!50,fill=blue!20,thick,minimum size=5mm},
entrada/.style={rectangle,draw=black!50,fill=black!20,thick,minimum size=5mm}]
\node[entrada] (w11)  at (0,10) {$u_1$};
\node[entrada] (w12) at (1,10) {$u_2$};
\node[entrada] (w13) at (2,10) {$u_3$};
\node[] (w14) at (4.5,9) {};
\node[entrada] (w15) at (5,10) {$u_i$};
\node[] (w16) at (5.5,9) {};
\node[entrada] (w17) at (8,10) {$u_{N}$};
\node[] (w17i) at (8,9) {};
\node[] (w17d) at (10,10) {Step 0};

\draw[-,dashed] (w13) -- (w15) {};
\draw[-,dashed] (w15) -- (w17) {};

\node[reservoir] (w21)  at (0,8) {$x_1$};
\node[reservoir] (w22) at (1,8) {$x_2$};
\node[reservoir] (w23) at (2,8) {$x_3$};
\node[reservoir] (w25i) at (3.9,8) {$x_{i-1}$};
\node[reservoir] (w25) at (5,8) {$x_i$};
\node[reservoir] (w25d) at (6.1,8) {$x_{i+1}$};
\node[reservoir] (w27) at (8,8) {$x_{N}$};
\node[] (w27d) at (10,8) {Step 1};
\node (w26) at (7.5,7) {};

\draw[-,dashed] (w23) -- (w25i) {};
\draw[-,dashed] (w25d) -- (w27) {};

\draw [->,blue] (w11) -- (w21);
\draw [->,blue] (w12) -- (w21);
\draw [->,blue] (w11) -- (w22);
\draw [->,blue] (w12) -- (w22);
\draw [->,blue] (w13) -- (w22);
\draw [->,blue] (w14) -- (w25);
\draw [->,blue] (w15) -- (w25);
\draw [->,blue] (w16) -- (w25);
\draw [->,blue] (w17) -- (w27);
\draw [->,blue] (w17i) -- (w27);

\node[reservoir] (w31)  at (0,6) {$x_1$};
\node[reservoir] (w32) at (1,6) {$x_2$};
\node[reservoir] (w35) at (5,6) {$x_i$};
\node[reservoir] (w37) at (8,6) {$x_{N}$};
\node[] (w37d) at (10,6) {Step 2};
\draw[-,dashed] (w32) -- (w35) {};
\draw[-,dashed] (w35) -- (w37) {};

\draw [->,blue] (w21) -- (w31);
\draw [->,blue] (w22) -- (w31);
\draw [->,blue] (w21) -- (w32);
\draw [->,blue] (w22) -- (w32);
\draw [->,blue] (w23) -- (w32);
\draw [->,blue] (w25i) -- (w35);
\draw [->,blue] (w25d) -- (w35);
\draw [->,blue] (w25) -- (w35);
\draw [->,blue] (w27) -- (w37);
\draw [->,blue] (w26) -- (w37);

\node[reservoir] (wm1)  at (0,3) {$x_1$};
\node[reservoir] (wm2) at (1,3) {$x_2$};
\node[reservoir] (wm5) at (5,3) {$x_i$};
\node[reservoir] (wm7) at (8,3) {$x_{N}$};
\node[] (wmd7) at (10,3) {Step M-1};

\draw[-,dashed] (wm2) -- (wm5) {};
\draw[-,dashed] (wm5) -- (wm7) {};
\draw[-,dashed] (w31) -- (wm1) {};
\draw[-,dashed] (w32) -- (wm2) {};
\draw[-,dashed] (w35) -- (wm5) {};
\draw[-,dashed] (w37) -- (wm7) {};


\end{tikzpicture}
\caption{Reservoir computing scheme using cellular automata. The input is stored in the reservoir at the beginning of the computation and evolved in time during different time steps. }
\label{ReCA}
\end{figure}
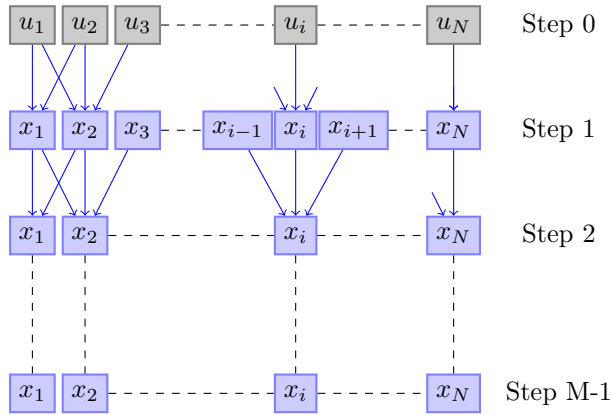
	
	The softmax regression has three main advantages. (1) It can handle multiclass classification problems. (2) Once the model is trained, it suffices to compute $\boldsymbol{\hat{Y}}$ to make a prediction, since the most likely categories are $\textnormal{argmax}(\boldsymbol{\hat{Y}}_{i})$. (3) The loss function \eqref{eq:lossfunction} is convex \cite{nasrabadi2007pattern}, i.e. it has only one global minimum.

\subsection{Fundamental behavior of ReCA systems}
A ReCA system is obtained when the state of each internal node of the reservoir is binary (zero or one) and the dynamics of the reservoir obeys a Cellular Automata rule. For Elementary Cellular Automata (ECA),  equation (\ref{dynamics}) is transformed to:

\begin{equation}
x_i(k)=F\Big[ x_{i-1}(k-1), x_{i}(k-1), x_{i+1}(k-1) \Big]
\end{equation}

where $F$ is a boolean function and $x_i \in \{0,1\}$. Therefore, as depicted in figure \ref{ReCA}, the state of the $ith$ cell depends on the  state of its neighbors ($x_{i-1}$ and $x_{i+1}$) and itself ($x_i$) evaluated in the previous time step for all the binary cells. The reservoir state $\boldsymbol{x}$ used to evaluate $\boldsymbol{\hat{y}}$ with equation \eqref{eq2} is therefore constructed from the automaton evolution for $M$ time steps $\boldsymbol{x} = \left[ \boldsymbol{x}_1\boldsymbol{(0)}, \boldsymbol{x}_2\boldsymbol{(0)}, \dots, \boldsymbol{x}_N\boldsymbol{(0)}, \boldsymbol{x}_1\boldsymbol{(1)}, \dots, \boldsymbol{x}_N\boldsymbol{(M)} \right]$.

  \begin{figure}[!ht]
  \centering
  \includegraphics[width=0.8\columnwidth]{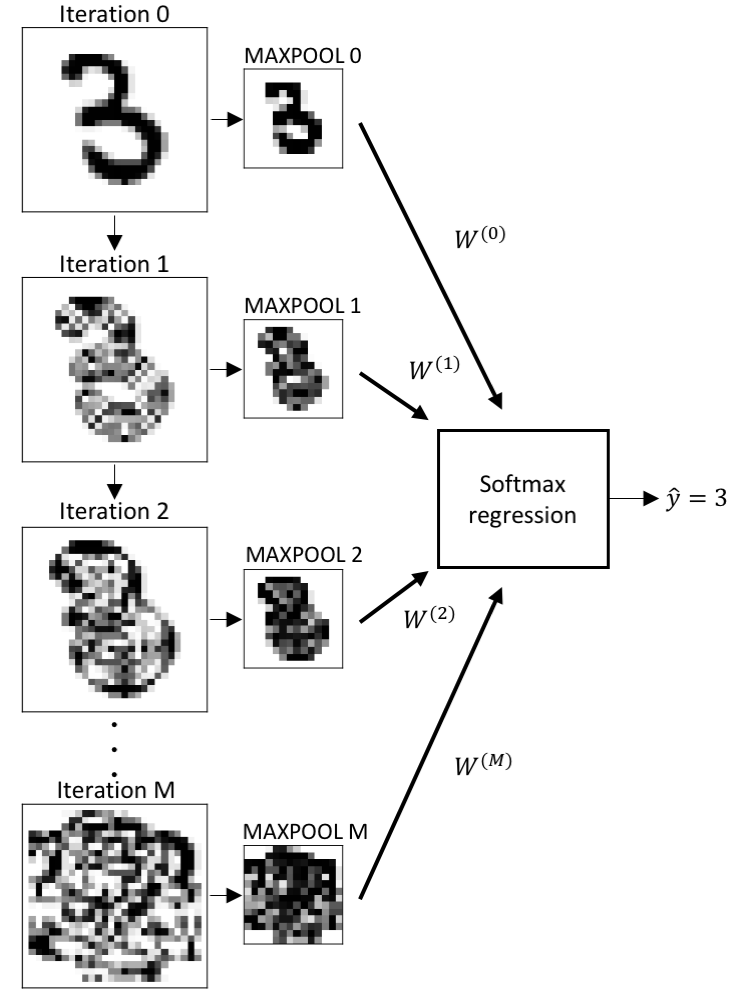}
  \caption{Scheme of the proposed classifier applied to a MNIST sample.}
  \label{fig:elm_ca_scheme}
  \end{figure}

\begin{figure}[!ht]
  	\centering
  	\includegraphics[width=0.7\columnwidth]{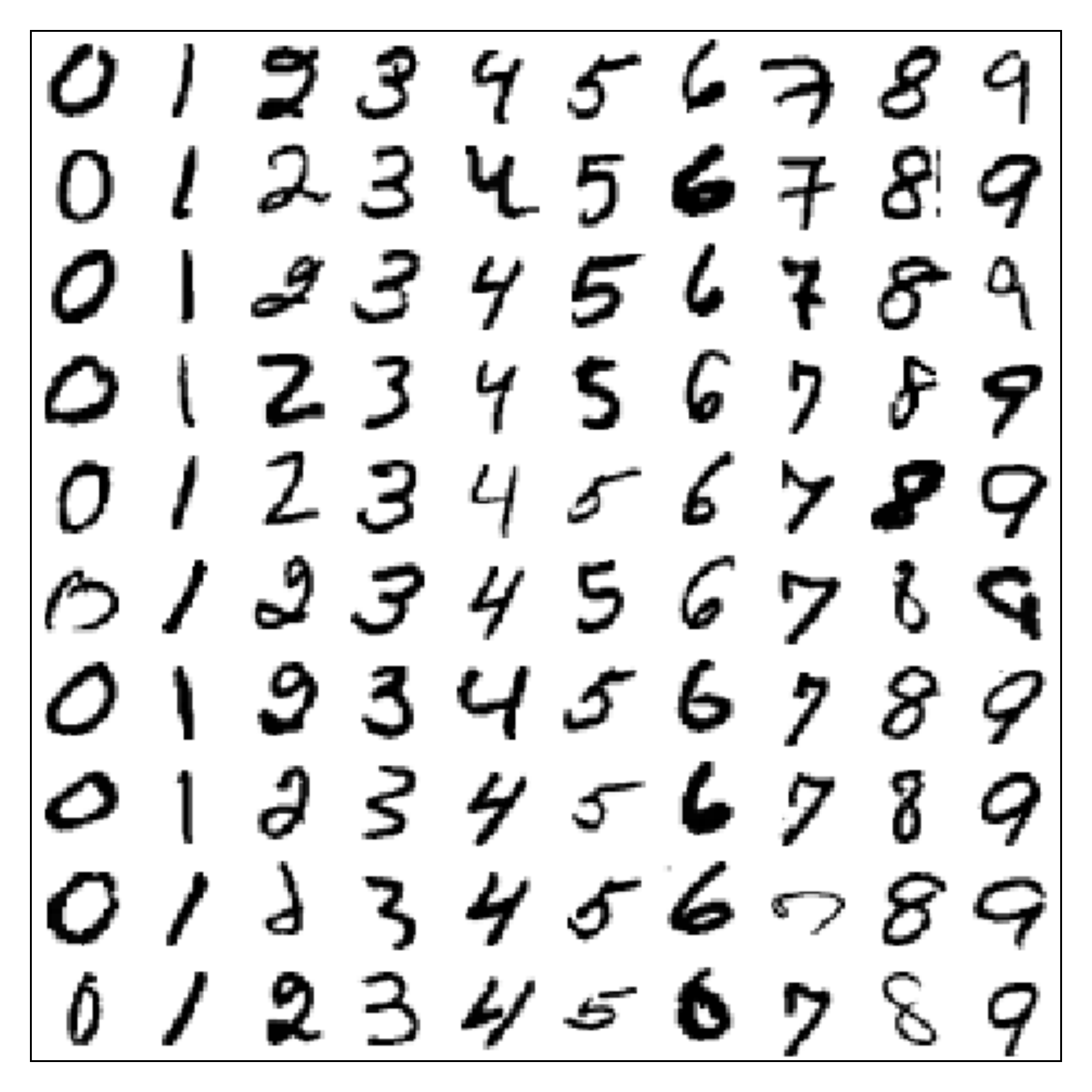}
  	\caption{Some samples taken from the MNIST database.}
  	\label{fig:mnistExamples}
  	\end{figure} 
  	
  \begin{figure*}[!ht]
  \centering
  \includegraphics[width=\columnwidth]{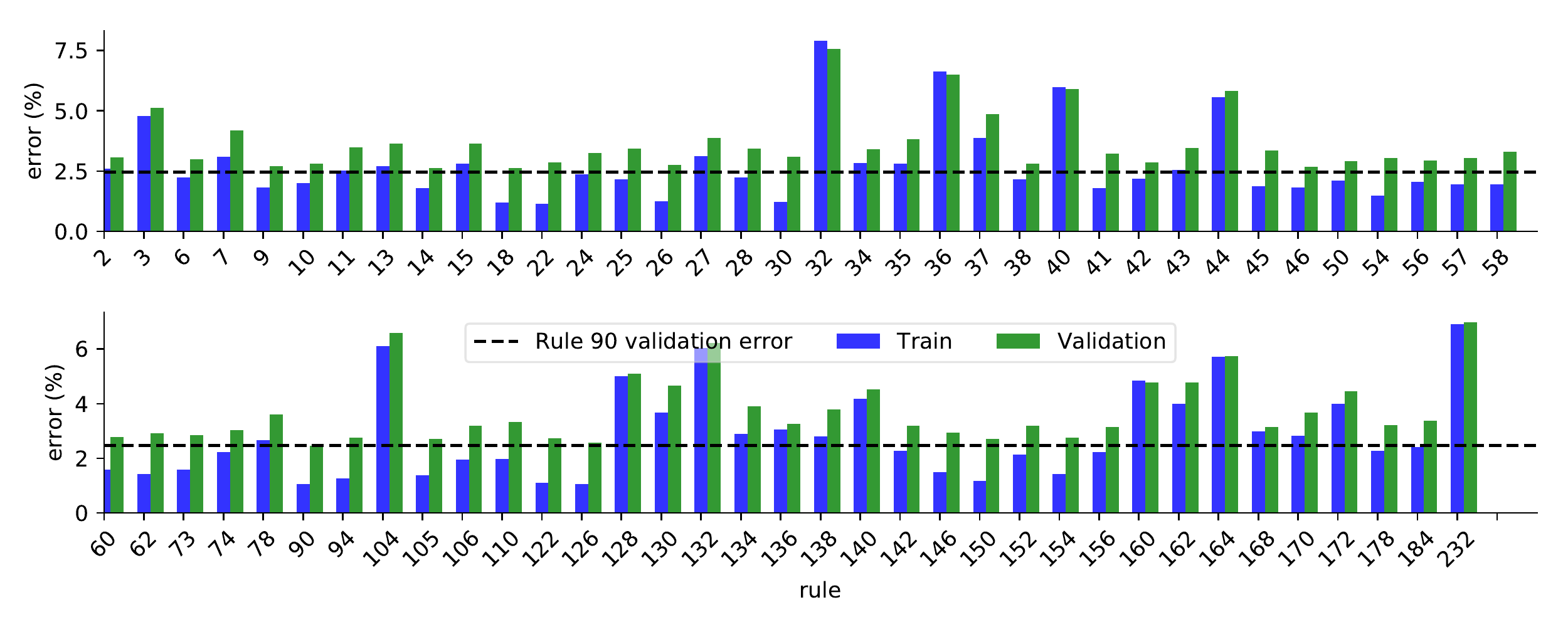}
  \caption{Performance of the proposed architecture using 10 iterations of different, non-symmetric ECA rules. ECA rules with vanishing or period 2 and temporal evolution without transient state are not shown in this figure.}
  \label{fig:error_several_rules}
  \end{figure*}   
 
\subsection{ReCA systems applied to image classification}
\label{subsec:reca}

For the special case of Reservoir Computing implemented using the dynamic of Elementary Cellular Automata (ReCA systems restricted to ECA), the intrinsic one-dimensional structure with binary states of ECA systems has to be adapted to higher dimensions. In the case of grayscale images, we deal with two spatial dimensions  where each pixel is characterized with a 8 bit signal value. In the case of RGB images, we should deal with three dimensions since each color channel (red, green and blue) has its own layer of intensities. So, the first question we address in this section is how do we extrapolate ECA, (optimum to iterate a 1-dimensional binary array) to process a higher dimensional signal.
  
For 2D grayscale images we iterate rows and columns independently using ECA rules and combine the resulting vectors using a bitwise XOR operation. The image ($\boldsymbol{u}$) is decomposed in 8 layers corresponding to each binary digit of each pixel:
\begin{equation}
\boldsymbol{u}=\sum_{j=0}^{B-1} 2^{j} \boldsymbol{u^{(j)}}.
\label{image_i}
 \end{equation}
\noindent where $\boldsymbol{u}^{(B-1)}$ and $\boldsymbol{u}^{(0)}$ are the most and less significative bits of $\boldsymbol{u}$ respectively. 
We construct a total of $B$ layers of reservoir automata $\boldsymbol{x(k)}$ that is evolved in time so that $\boldsymbol{x(0)}=\boldsymbol{u}$ and $\boldsymbol{x(k)}$ is defined as
 \begin{equation}
 \boldsymbol{x(k)} = \sum_{j=0}^{B-1} 2^{j} \boldsymbol{x^{(j)}(k)}.
 \end{equation}
So that $\boldsymbol{x^{(j)}(k)}$ is a boolean time-dependent image with initial value $\boldsymbol{x^{(j)}(0)}=\boldsymbol{u^{(j)}}$ and whose dynamics is governed by a cellular automata. Therefore, the automata rules are applied to each layer, with no communication between layers. The dynamic behavior of the classifier is illustrated in Fig.\ref{fig:elm_ca_scheme} using a MNIST sample. As can be seen, the input image $\boldsymbol{u}$ is evolved in time during a total of M time steps (thus creating signal $\boldsymbol{x(k)}$). Note that in the particular case of grayscale images, the binary tensor has $3$ dimensions (the width, $w$, the height, $h$, and the depth, $B$).
  
   Rows and columns are iterated independently applying the same automaton rule repeatedly for each iteration, starting from the input image $\boldsymbol{u}\in \mathbb{N}^{w\times h}$. Let $g^k$ be the function $g$ applied $k$ times, e.g. $g^3(\cdot)=g(g(g(\cdot)))$. Then, the input iterated by rows is given by
   \begin{equation}\label{eq:rows}
   \begin{split}
  	\boldsymbol{x_{r}^{(l)}(k)} &= g_{r}^{k}(\boldsymbol{u^{(l)}})\\
  	&= \left\{
\begin{array}{c l}
 \boldsymbol{u^{(l)}} & k=0\\
 g_{r}\left( \boldsymbol{x_r^{(l)}(k-1)} \right) & k>0
\end{array}
\right. \hspace{0.2cm} \forall l \in [0,B-1]
  \end{split}
  \end{equation}
  where the function $g_{r}: \{0,1\}^{w\times h} \rightarrow \{0,1\}^{w\times h}$ iterates the rows of the binary layers of the input from the most to the less significative digits. Similarly, the columns' iterations of the input are given by
     \begin{equation}\label{eq:columns}
   \begin{split}
  	\boldsymbol{x_{c}^{(l)}(k)} &= g_{c}^{k}(\boldsymbol{u^{(l)}})\\
  	&= \left\{
\begin{array}{c l}
 \boldsymbol{u^{(l)}} & k=0\\
 g_{c}\left( \boldsymbol{x_c^{(l)}(k-1)} \right) & k>0
\end{array}
\right. \hspace{0.2cm} \forall l \in [0,B-1]
  \end{split}
  \end{equation}
  with $g_{c}: \{0,1\}^{w\times h} \rightarrow \{0,1\}^{w\times h}$.
  
  Finally, these contributions obtained independently $\left[\boldsymbol{x_r(k)},\boldsymbol{x_c(k)}\right]$ are combined using a XOR function
    \begin{equation}\label{eq:combine_rows_and_columns}
    \begin{split}
  	\boldsymbol{x^{(l)}(k)} = 
 \boldsymbol{x_r^{(l)}(k)} \oplus \boldsymbol{x_c^{(l)}(k)}
 \hspace{0.2cm}\forall l \in [0,B-1],
  \end{split}
  \end{equation}
  where $\oplus$ represents a bitwise exclussive-OR operation that is restricted to the intersection bit between the row and column vector.
  
 The method can be generalized to higher dimensions (e.g. in the case of RGB images), following the same convention used for rows and columns in equations \eqref{eq:rows} and \eqref{eq:columns}, and then combining all the contributions using a single $\textnormal{XOR}$ to obtain $\boldsymbol{x^{(l)}}$, as in equation \eqref{eq:combine_rows_and_columns}.
  
  To obtain invariance under small translations and reduce the number of weights in the output layer, we apply max pooling filters to the original image and after each iteration (illustrated in Fig.\ref{fig:elm_ca_scheme}). Pooling layers are selected to have a stride of 2, zero padding and squared window of size 2.
  
  In contrast to the evolution of the automaton iterated by rows and columns, max pooling is applied to each iteration. Therefore, after applying max pooling and considering that $w$ and $h$ are multiples of 2, (as is the case of the MNIST digits) then the input to the softmax classifier for $M$ iterations is $\boldsymbol{x_f}=\textnormal{maxpool}\left( \boldsymbol{x} \right)\in \mathbb{N}^{\frac{w}{2}\times\frac{h}{2}\times M}$ rearranged as a single column vector $\boldsymbol{x_{f}}$, which is referred to as the feature vector. Therefore, the total number of weights used by the model is $\frac{Mwh}{4}$. Ideally, the feature vector must not include redundant information, so that a rich non-linear behaviour to obtain a good classification accuracy is necessary.  	
  
  \section{Results}
  \label{sec:results}
  The study of the software and hadware implementations of the ReCA system have been done with the MNIST database and is divided in two parts. First, we use a double precision software simulation from which we select the ECA rule with better results. Next, the selected classifier is trained also by software using 8-bits precision weights that are used in a final FPGA implementation. The training is carried out using mini batches and Adam optimization method, reaching equivalent performance than with double precision weights.

  	The MNIST database consists of 70000 squared, grayscale digit images of $28\times 28$ pixels, some examples are shown in Fig.\ref{fig:mnistExamples}. These 70000 images are divided in two sets: 60000 for optimization and 10000 for testing. The model is adjusted using a training and a validation set. In our case, 55000 samples of the first set are used to optimize the weights, while the other 5000 samples are used as validation set to tune the hyperparameters (e.g. automaton rule, iterations and regularization parameter).
  
  \subsection{Software exploration}
  \label{subsec:double} 
Using the reservoir methodology described in section \ref{subsec:reca} we check the performance of all the ECA rules in the processing of the MNIST database. In Fig.\ref{fig:error_several_rules} we show the results obtained for all the ECA rules that do not share any symmetry. Also, trivial rules with vanishing or period two and without transient state from random initial conditions are not shown since they give worse results than the ones presented in this figure. From these results we select rule 90 since it is providing one of the lowest errors and a simple digital implementation.
  
    \begin{figure}[ht!]
    \centering
    \includegraphics[width=0.6\columnwidth]{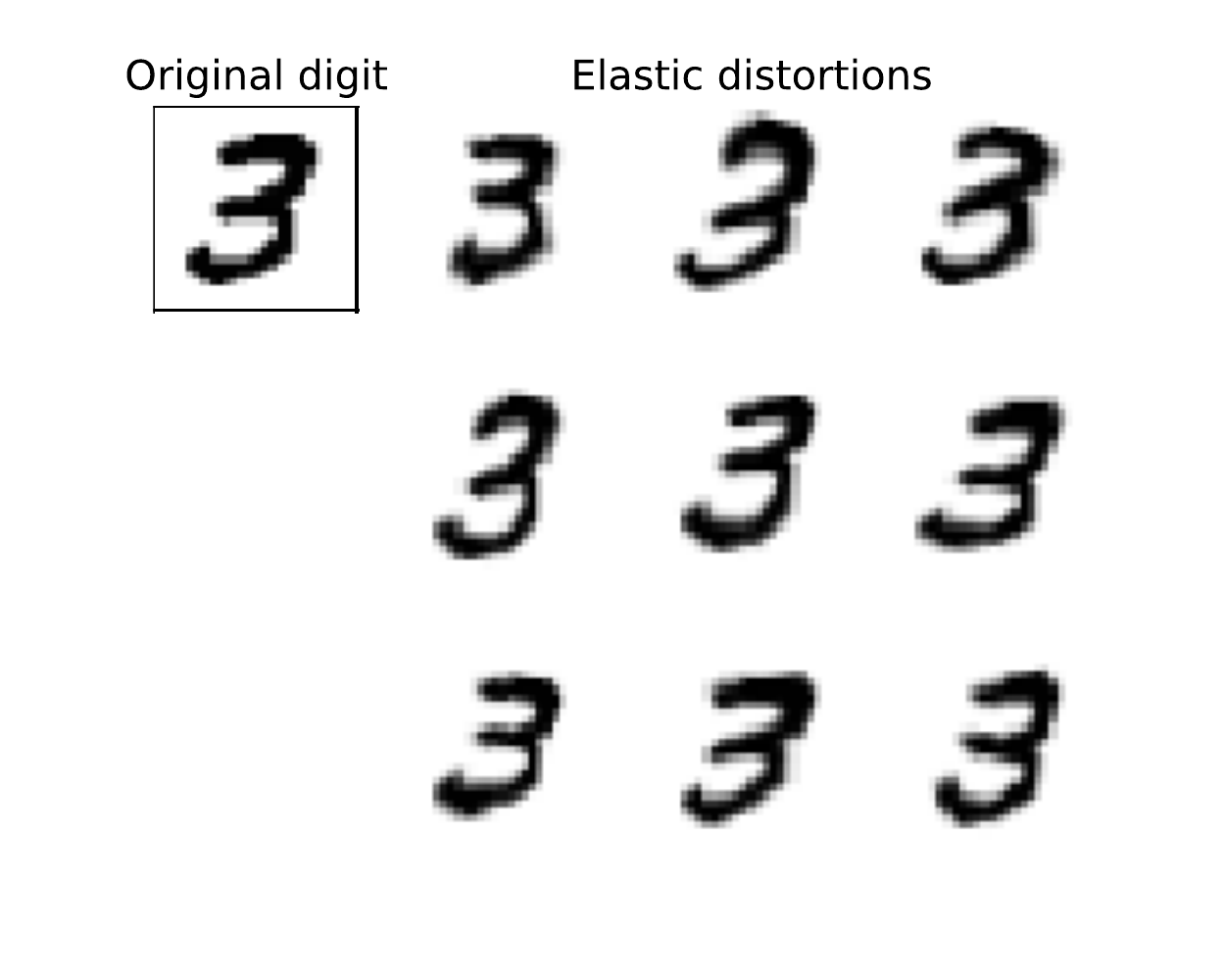}
    \caption{Example of 9 elastic distortions applied to a sample MNIST digit using $\alpha_d=30$ and $\sigma_d=5$.}
    \label{fig:elasticDistortions}
    \end{figure}
  
 To improve the training performance, we expand the training set through elastic distortions \cite{simard2003best}. An example is shown in Fig.\ref{fig:elasticDistortions} using two parameters. The first ($\alpha_d$) quantifies the amount of displacement per pixel, and the second ($\sigma_d$) the standard deviation of a Gaussian kernel that is applied to convolve the image.
 
  \begin{figure}[!ht]
  \centering
  \includegraphics[width=0.9\columnwidth]{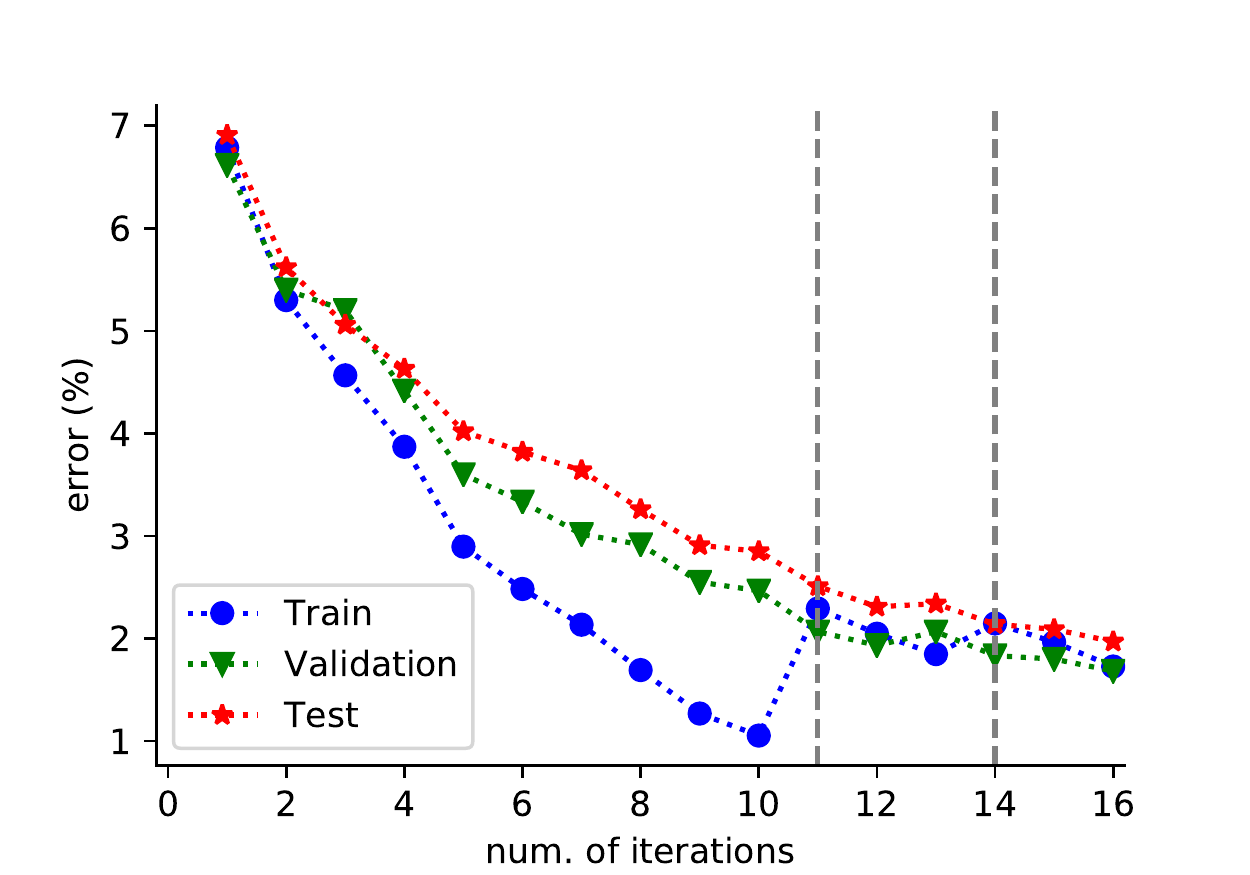}
    \caption{MNIST training, validation and test error, using rule 90 as described in section \ref{subsec:reca}, adding 55000 additional samples using elastic distortions every vertical dashed line (at $M=11$ and $M=14$). The regularization parameter is $C=0.04$ and we fix the maximum number of limited memory BFGS iterations to 1000.}
  \label{fig:error_rule90}
  \end{figure}

  In Fig.\ref{fig:error_rule90} we show the error obtained as the size of the reservoir is increased (number of iterations). Also, the training set is augmented twice when $M=11$ and $M=14$. When the gap between the training and validation errors becomes large enough and results cannot be further improved via a higher regularization, we add elastic distortions (each training set increment is indicated with a vertical dashed line in Fig.\ref{fig:error_rule90}). The training stage has been performed using 55000 samples for $M<11$, 110000 samples for $11\leq M <14$ and 165000 for $M\geq 14$.  
  
  \subsection{Hardware implementation of the proposed model}    
  For the hardware implementation of the ReCA system, a 8-bits weights model is used to obtain similar results than with double precision. 
  
  The training is done using Adam \cite{kingma2014adam}, which is an algorithm for first-order gradient based stochastic optimization, applied to equation \eqref{eq:lossfunction} in our case. On the one hand, this algorithm is first-order because, unlike the limited memory BFGS method (in which an approximation of the Hessian matrix is computed) only the gradient and a few additional calculations are needed to update the weights. The gradient of the loss function \eqref{eq:lossfunction} is computed using a randomly selected fixed-size subset (or mini batch) taken from the training set. The weights are updated and the mini-batch changed each training step.
  
  The loss function  expressed in \eqref{eq:lossfunction} ($\mathcal{L}(\boldsymbol{W})$ for short) is a noisy objective function when using batches. During the training process this function changes since both the input data and the parameters change. To indicate this dependence with time step $t$, we use $\mathcal{L}_t(\boldsymbol{W}_t)$ instead, so that the gradients are $\boldsymbol{g}_t=\nabla_{\boldsymbol{W}}\mathcal{L}_t(\boldsymbol{W}_t)$. Since the objective is to minimize the expected value $\mathbb{E}[\mathcal{L}_t(\boldsymbol{W}_t)]$, the algorithm updates exponential moving averages of  $\boldsymbol{g_t}$ and $\boldsymbol{g}_t^2$, which are refered to as $\boldsymbol{m}_t$ and $\boldsymbol{v}_t$ respectively. To control the decay rates of these two exponential moving averages, two new hyper-parameters are introduced: $\beta_1, \beta_2\in[0,1)$, in addition to the learning rate $\alpha$. 
  
The algorithm is implemented as follows. At the beginning ($t=0$), both moving averages are initialized to zero, i.e. $\boldsymbol{m}_0=\boldsymbol{v}_0=0$, and the first mini batch is selected. Then, for each new training iteration, the gradients $\boldsymbol{g}_t$ are computed to update the moving averages as
\begin{equation}
	\boldsymbol{m}_t = \frac{\beta_1 \boldsymbol{m}_{t-1} + (1-\beta_1)\boldsymbol{g}_t}{1-\beta_1^t},
\end{equation}
\begin{equation}
	\boldsymbol{v}_t = \frac{\beta_2 \boldsymbol{v}_{t-1} + (1-\beta_2)\boldsymbol{g}_t^2}{1-\beta_2^t}.
\end{equation}
Finally, the weights are updated as
\begin{equation}
	\boldsymbol{W}_{t} = \boldsymbol{W}_{t-1} - \alpha\frac{\boldsymbol{m}_t}{\sqrt{\boldsymbol{v}_t}}
\end{equation}
and the process is repeated using a new mini batch until convergence of $\boldsymbol{W}_t$. Note that the term $\boldsymbol{m}_t/\sqrt{\boldsymbol{v}_t}$ is a smoother version of the gradient that depends on previous computations. Thus, the variance of the stochastic trajectory of $\mathcal{L}_t(\boldsymbol{W}_t)$ becomes smaller compared to simple stochastic gradient descent, which also results in faster convergence.

  \begin{table}[!ht]
  \caption{This table specifies the hyperparameters of the model with 8-bits weights. (*) We used the Tensorflow \cite{abadi2016tensorflow} AdamOptimizer class. (**) Mini-batches are chosen at random for each optimization step.}
\label{tab:hyperparameters}
\centering
\begin{tabular}{lc}
\multicolumn{2}{c}{Hyperparameters of the quantized model}                                                                                                                          \\[1ex] \hline
\multicolumn{1}{|l|}{CA rule}                                                                                      & \multicolumn{1}{c|}{90}                                        \\ \hline
\multicolumn{1}{|l|}{CA iterations ($M$)}                                                                                & \multicolumn{1}{c|}{16}                                        \\ \hline
\multicolumn{1}{|l|}{Optimization method}                                                                          & \multicolumn{1}{c|}{Adam* ($\beta_1 = 0.9$, $\beta_2 = 0.999$)} \\ \hline
\multicolumn{1}{|l|}{Learning rate}                                                                                & \multicolumn{1}{c|}{0.008}                                     \\ \hline
\multicolumn{1}{|l|}{Regularization strength}                                                                      & \multicolumn{1}{c|}{0.00012}                                  \\ \hline
\multicolumn{1}{|l|}{Batch size**}                                                                                  & \multicolumn{1}{c|}{17000}                                      \\ \hline
\multicolumn{1}{|l|}{\begin{tabular}[c]{@{}l@{}}Distortions per image\\ ($\alpha_d = 30$, $\sigma_d = 5$)\end{tabular}} & \multicolumn{1}{c|}{3}                                         \\ \hline
\end{tabular}
\end{table}

  \begin{figure}[!ht]
  \centering
  \includegraphics[width=0.9\columnwidth]{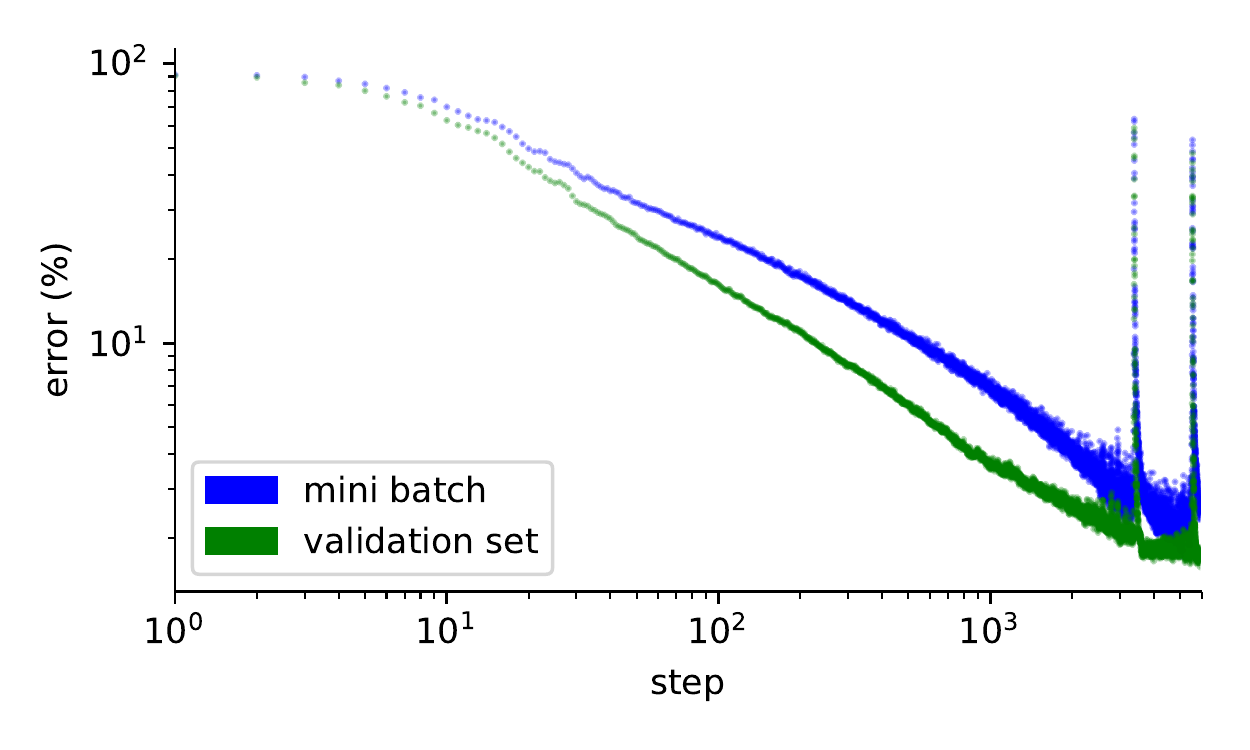}
  \caption{Training process for the quantized model using the hyperparameters specified in table \ref{tab:hyperparameters}.}
  \label{fig:training_process}
  \end{figure}
%
%
\begin{figure*}[!ht]
\centering
\includegraphics[width=\columnwidth]{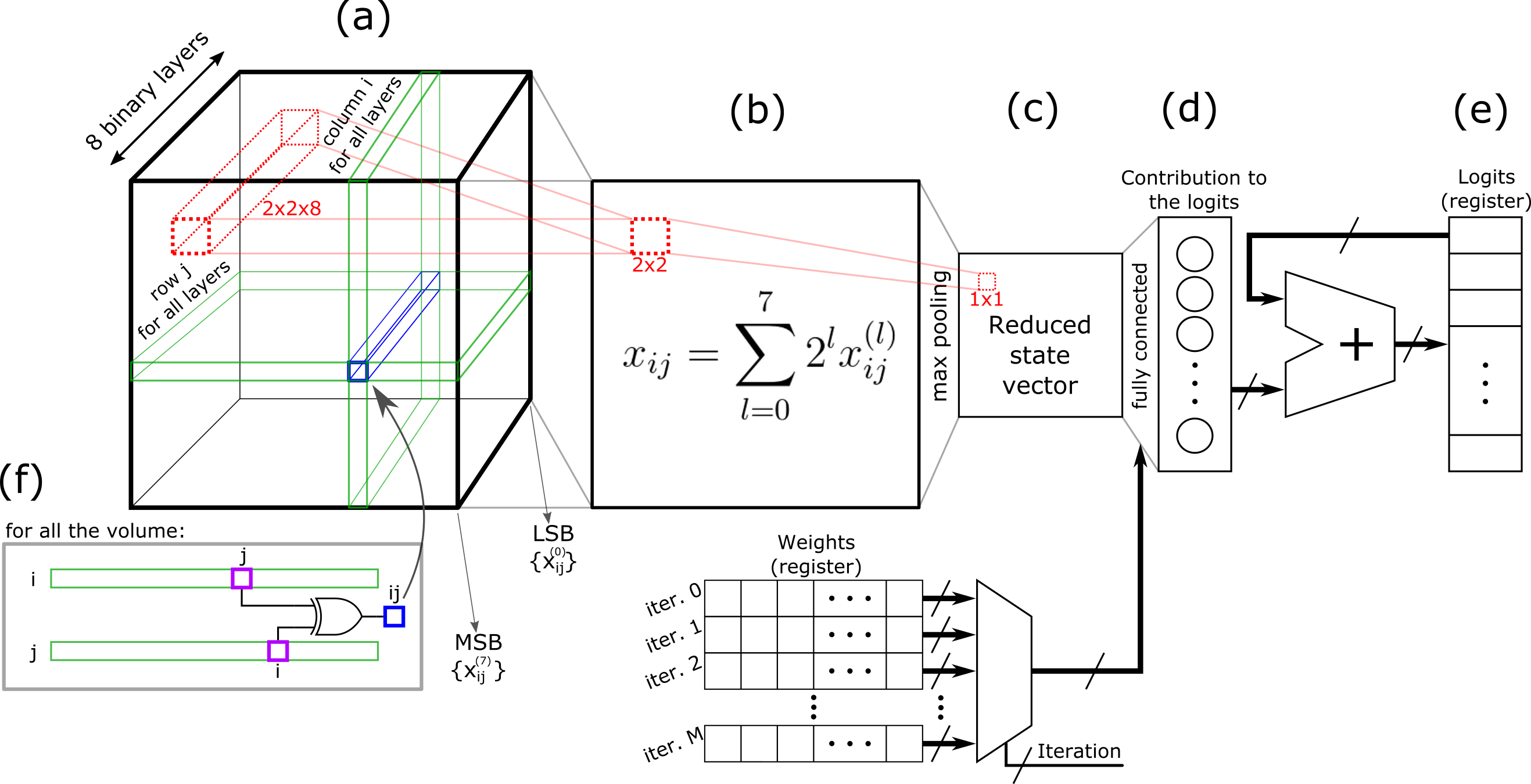}
\caption{	An scheme of the hardware implementation of the proposed classifier.(a) 3D boolean tensor representation of 2D grayscale data, this tensor representation is organized in 2D layers from the most significative bits (MSB) to the less significative bits (LSB) of the grayscale image. (b) The 8 binary layers are interpreted as unsigned integers $x_{ij}$. (c) Max pooling filter applied to $\boldsymbol{x}$. (d) The contribution to the logits is computed using the pre-stored weights obtained offline. (e) The current contribution to the logits is accumulated until the final iteration is reached. (f) The rows and columns are iterated independently using rule 90 and combined to obtain an updated 3D boolean tensor using a bitwise XOR operation. This process is repeated until the final iteration is reached. Once the contibution of the final iteration is accumulated, the logits are valid data.}
\label{fig:tensor_ecasoftmax}
\end{figure*}
  
  Rule 90 and a total of 16 iterations are implemented in the hardware (all the parameters are listed in table \ref{tab:hyperparameters}). The training process using these parameters and Adam optimization method is illustrated in Fig.\ref{fig:training_process}. In this figure we represent the error in each mini-batch (blue) and the error in the validation set (green) as a function of the number of optimization steps, i.e. the number of times the optimization algorithm has been applied. After a large number of optimization steps, we observed peaks of poor performance due to 8-bits quantization restrictions, which slows down the training process. The training ends when the validation error is less than $1.6\%$. As a result we obtained a $1.92\%$ of error in the test set.
  
  The proposed method has been implemented in hardware using weights estimated offline. In Fig.\ref{fig:tensor_ecasoftmax} we show the general scheme of the implemented circuitry. Some additional details of the FPGA implementation are specified in table \ref{tab:fpga_details}. We used a total of 40 DSP blocks, 4 of them for each category, so that the contribution to the matrix multiplication of each iteration is parallelized across 4 DSP blocks for each output. Note that this choice is quite arbitrary and the number of DSP blocks is limited by the specific device. Depending on the dimensions of the input data and the number of required iterations for a specific task, the number of DSP blocks could be chosen to obtain a fast classification speed.

\begin{figure}[!ht]
\centering
\includegraphics[width=0.8\columnwidth]{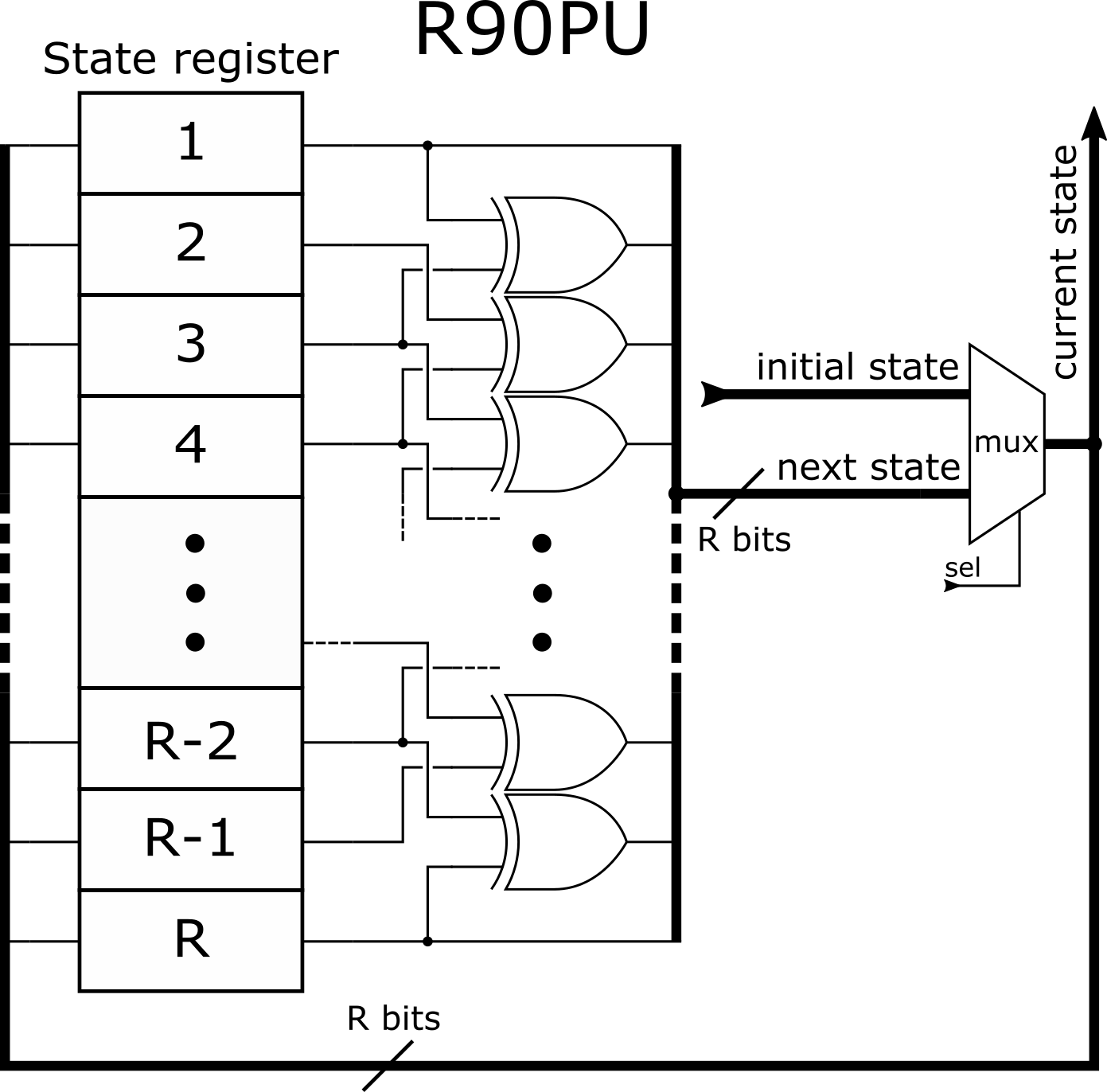}
\caption{Digital scheme of the rule 90 processing unit (R90PU). The present state is indexed from 1 to $R$ ($R=28$ in the case of the MNIST digits) and rule 90 is implemented by computing the XOR of the nearest neighbours, except for the first and last elements, for which we consider fixed boundary conditions.}
\label{fig:r90pu-ecasoftmax}
\end{figure}

The core of the ReCA processing is shown in Fig.\ref{fig:r90pu-ecasoftmax}. The ECA rule selected (90) is not only the one with the lowest prediction error, but also one of the most simplest to be implemented since only one XOR gate is needed to implement the rule. The rule 90 processing unit (R90PU) depicted in Fig.\ref{fig:r90pu-ecasoftmax} is used for each row and column, which are iterated independently. Therefore, in the case of the MNIST, we require a total of $2 \times 8 \times 28 = 448$ R90PUs with 28 registers ($R=28$) each one. Additionally, a total of $8 \times 28\times 28 = 6272$ XOR gates are needed to account for the bitwise XOR of the iteration computed by rows and columns in parallel (equation \eqref{eq:combine_rows_and_columns}). In Fig.\ref{fig:tensor_ecasoftmax}, two R90PUs layers are highlighted in green color, that provide all the bits ($x_{ij}^{(l)}$) needed to reproduce pixel $X_{ij}$ (in blue). For each $2\times 2$ pixels evaluated (Fig.\ref{fig:tensor_ecasoftmax}b in red), the one with the highest value is selected (Fig.\ref{fig:tensor_ecasoftmax}c). This reduced max pooling layer is multiplied by its specific weight depending on the specific iteration (provided by the weights register). All these computations are sequentially accumulated in the output composed by $Q=10$ registers.

   \begin{table}[!ht]
   \caption{FPGA implementation details of the proposed classifier using rule 90, 16 iterations and 8-bits precision weights. (*) The power consumption has been calculated using the \textit{PowerPlay Early Power Estimator} provided by Altera \cite{quartus2013handbook}. The FPGA device is an \textit{Altera Cyclone V 5CSEBA6U23I7 SoC}.}
\label{tab:fpga_details}
\centering
\begin{tabular}{lcc}
\cline{2-3} 
\multicolumn{1}{l|}{}                          & \multicolumn{1}{l|}{\textbf{Test bench}} & \multicolumn{1}{l|}{\textbf{Classifier}} \\ \hline
\multicolumn{1}{|l|}{Logic utilization (ALMs)} & \multicolumn{1}{c|}{25.5K ($23.2\%$)}                      & \multicolumn{1}{c|}{22.6K ($20.5\%$)}                \\ \hline
\multicolumn{1}{|l|}{DSP blocks}               & \multicolumn{1}{c|}{40}                       & \multicolumn{1}{c|}{40}                  \\ \hline
\multicolumn{1}{|l|}{FPGA power consumption*}  & \multicolumn{1}{c|}{0.289 W}                  & \multicolumn{1}{c|}{-}               \\ \hline
\end{tabular}
\end{table}
    
  Additionally, we compare our implementation to previous publications implementing popular models of CNNs in terms of accuracy, latency and resource usage (see table \ref{tab:comparison}). From this table we can appreciate that our model is faster than conventional CNNs and can provide good accuracy while maintaining a relatively low power consumption.
  \begin{table*}[!ht]
  \caption{FPGA implementation of the proposed model using 8-bits precission weights compared to some previous works.}
\label{tab:comparison}
  \scriptsize
\centering
\resizebox{\columnwidth}{!}{
\begin{tabular}{llllllll}
Model     & \thead{  Accuracy \\($\%$)} & \thead{ Power \\(W)} & \thead{Latency \\(ms)}  & \thead{Power-delay\\ product (mJ)} & \thead{Clock freq.\\ (MHz)} & Gate size    &  \thead{ DSP \\blocks} \\[1px] \hline \\[-3px]
This work & 98.08    & 0.289     & 0.020         &  0.00578   & 50                    & 22.6K ALM            & 40         \\
\cite{liang2018fp}       & 98.32    & 26.2      & 0.0034          &  0.0891  & 150                   & 182.3K ALM           & 20         \\
\cite{lin2017binarized}       & 99.52    & -                 & 4-6 (approx.) & -   & -                     & 36.4K LUT + 41.1K FF & 8          \\
\cite{ghaffari2016fpga}       & 98.62    & -                 & 26.37         & -   & 100                   & 14.8K LUT + 54.1K FF & 20         \\
\cite{zhou2015fpga}       & 96.8     & -                 & 0.0254        & -   & 150                   & 51.1K LUT + 66.3K FF & 638       
\end{tabular}
}	
\end{table*}
   
  \section{Conclusion}
  \label{sec:conclusion} 
  Due to the discrete nature of Cellular Automata, we note that our classifier is specially attractive to handle integer valued data (e.g. grayscale images) but it could also be used with floating point data using two's complement convention to iterate the CA, as is done in \cite{Nichele2017deep}. 
  
  The proposed ReCA system is implemented and tested using the MNIST data set, obtaining a fairly good accuracy, low latency, a reasonable number of logic elements and a low power consumption compared to previous works. Nevertheless, there is a tradeoff between classification speed (number of CA iterations) and accuracy, which may increase the latency when dealing with more complex data sets.
  
  Although the obtained accuracy is still not as good as that from state of the art models, our classifier based on ECA reservoir, max pooling and softmax regression represents a proof of concept for the applicability of ReCA systems for image classification. Moreover, its hardware implementation represents a fast and efficient alternative compared with previously-published hardware implementations of image recognition systems of CNN popular architectures.

  Further work in this line might consist of a similar hardware implementations using diferent image data sets (e.g. CIFAR-10 \cite{krizhevsky2009learning} or ImageNet \cite{Krizhevsky2012}) or time dependent data (e.g. ECG or EEG), considering a larger CA neighborhood, performing weight binarization \cite{rastegari2016xnor,courbariaux2016binarized} to further reduce the number of logic elements and DSP blocks and/or adding fully connected hidden layers before the softmax layer.

  \section*{Acknowledgments}

This work has been partially supported by the Spanish Ministry of Economy and Competitiveness (MINECO) and the Regional European Development Funds (FEDER) under grant contracts TEC2014-56244-R and TEC2017-84877-R.




%
  \bibliographystyle{ieeetr}
\bibliography{mybib2}

\end{document}